\pdfoutput=1
\documentclass[11pt]{article}
\usepackage{amsmath}
\usepackage{amssymb}
\usepackage[preprint]{acl}
\usepackage{multirow}
\usepackage{times}
\usepackage{latexsym}

\usepackage[T1]{fontenc}
\usepackage[utf8]{inputenc}

\usepackage{microtype}

\usepackage{inconsolata}

\usepackage{graphicx}

\title{TriSPrompt: A Hierarchical Soft Prompt Model for Multimodal Rumor Detection with Incomplete Modalities}

\author{Jiajun Chen$^1$, Yangyang Wu$^2$, Xiaoye Miao$^1$, Mengying Zhu$^2$, Meng Xi$^2$\\
$^1$ Center for Data Science, Zhejiang University, Hangzhou, China \\
$^2$ School of Software Technology, Zhejiang University, Ningbo, China \\
\texttt{\{chenjjcccc, zjuwuyy, miaoxy, mengyingzhu, ximeng\} @zju.edu.cn} \\}

\usepackage{booktabs}

\begin{document}
\maketitle
\begin{abstract}
 The widespread presence of incomplete modalities in multimodal data poses a significant challenge to achieving accurate rumor detection.
Existing multimodal rumor detection methods primarily focus on learning joint modality representations from \emph{complete} multimodal training data, rendering them ineffective in addressing the common occurrence of \emph{missing modalities} in real-world scenarios.
In this paper, we propose a hierarchical soft prompt model \textsf{TriSPrompt}, which integrates three types of prompts, \textit{i.e.}, \emph{modality-aware} (MA) prompt, \emph{modality-missing} (MM) prompt, and \emph{mutual-views} (MV) prompt, to effectively detect rumors in incomplete multimodal data.
The MA prompt captures both heterogeneous information from specific modalities and homogeneous features from available data, aiding in modality recovery. The MM prompt models missing states in incomplete data, enhancing the model's adaptability to missing information. The MV prompt learns relationships between subjective (\textit{i.e.}, text and image) and objective (\textit{i.e.}, comments) perspectives, effectively detecting rumors.
Extensive experiments on three real-world benchmarks demonstrate that \textsf{TriSPrompt} achieves an accuracy gain of over 13\% compared to state-of-the-art methods.
The codes and datasets are available at https: //anonymous.4open.science/r/code-3E88.
\end{abstract}

\section{Introduction}
Social media has emerged as a primary platform for the rapid dissemination of information, where multimodal communication—integrating various forms of media—has surpassed traditional unimodal formats to become the dominant mode \cite{wang2024modality}.
Compared to unimodal content, multimodal information conveys richer and more nuanced messages \cite{tandoc2018defining}, enhancing engagement but also introducing greater complexity, which poses significant challenges for effective rumor detection \cite{zou2024pvcg}.

\begin{figure}[t]
\centering
  \includegraphics[width=\columnwidth]{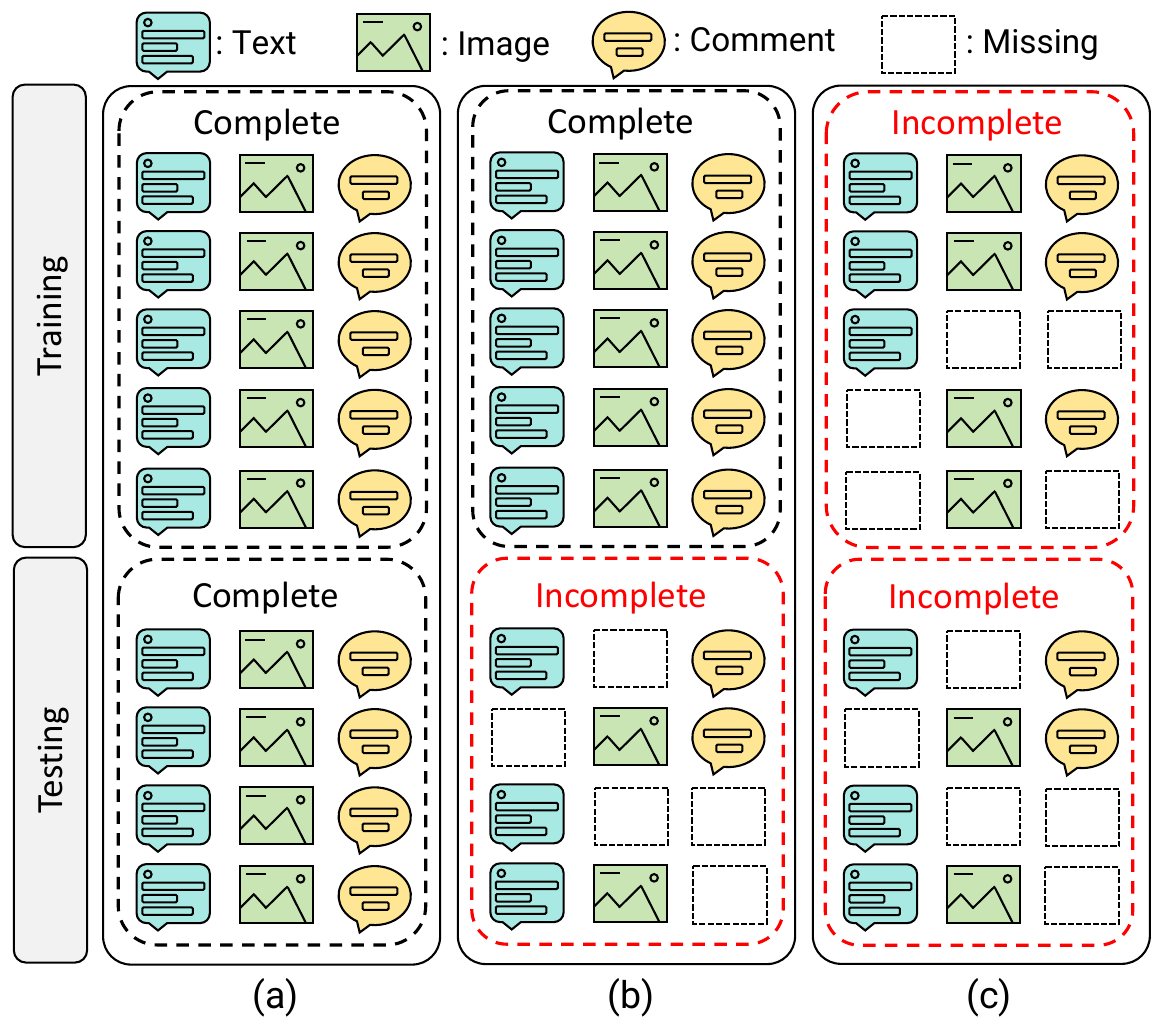}
  \vspace*{-0.25in}
  \caption{Multimodal rumor detection configurations.
(a) \emph{Cross-phase completeness alignment}: Full modalities in both phases (\textit{ideal data integrity scenario}).
(b) \emph{Cross-phase completeness discrepancy}: Train with full modalities; test with incomplete modalities.
(c) \emph{Cross-phase incompleteness consistency}: Incomplete modalities in both phases (\textit{real-world modality loss scenarios}).}
  % T (Text), I (Image), C (Comment).}
    % \caption{Existing methods cover only (a) full-modality training \& testing and (b) only testing with missing modalities. Our method tackles the most general case: (c) missing modalities in both training and testing.}
  \label{fig:configurations}
  \vspace*{-0.228in}
\end{figure}

Earlier research on rumor detection predominantly centered on text-based (unimodal) analysis \cite{rubin2015towards,przybyla2020capturing}, leveraging features such as writing style and dissemination patterns to identify rumors. 
However, as multimodal content has become the dominant form of communication, the focus has shifted towards multimodal rumor detection. This approach integrates information from different modalities, such as text and images, to enhance detection accuracy. For instance, some studies employ feature fusion techniques to combine text and image data, aiming to improve performance. Yet, simple fusion methods often fall short of capturing the intricate relationships between modalities. To overcome this limitation, \citet{wu2021multimodal} introduced a cross-attention mechanism to more effectively integrate features from each modality. Building on this, \citet{zheng2022mfan} utilized graph neural networks to incorporate contextual information from comments.
\citet{dhawan2024game} proposed a fine-grained fusion approach that directly operates on the original modality information using graph-based methods, representing a significant advancement in multimodal rumor detection research.
However, existing multimodal rumor detection methods often fail to address two critical practical challenges.

Firstly, \emph{incomplete modalities (\textbf{CH1}).}
In real-world scenarios, rumors spread rapidly, often outpacing the ability to collect complete multimodal data.
For instance, during the COVID-19 pandemic in 2020, a deluge of false information and rumors circulated rapidly, leading to inevitable modality loss during data collection due to factors such as network instability, device limitations, or other uncontrollable circumstances.
Despite this reality, most existing methods assume the availability of complete modalities. While CLKD-IMRD \cite{xu2023leveraging} partially addresses missing modalities using a distillation model, its reliance on complete training data significantly limits its generalizability. When faced with more common scenarios where both training and testing data are incomplete, as illustrated in Figure \ref{fig:configurations}, the performance of such methods degrades considerably.
Moreover, existing \emph{Multimodal Large Language Models} (MLLMs), excelling in understanding multimodal content, struggle with rumor detection due to their inherent assumption that input information is "reality" and lack a mechanism for verifying content accuracy.

%Firstly, \emph{incomplete modalities.} In the real world, rumors break out and spread rapidly like a hurricane. For example, during the COVID-19 pandemic in 2020, a significant amount of false information and rumors spread rapidly which leads to the inevitability of modality missing during the data collection process due to network, device and other uncontrollable factors. While, the existing methods are all proposed under the premise that modalities are complete. While CLKD-IMRD \cite{xu2023leveraging} addresses missing modalities through a distillation model, its reliance on complete training data limits its general applicability. In the event of encountering more common missing situations, namely instances where both the training data and the test data are incomplete as show in Figure \ref{fig:configurations}, these methods will be significantly affected, and we have verified this in our experiments.
Secondly, \emph{the nature of rumors (\textbf{CH2}).}
In a post comprising text, images, and comments, each modality plays a distinct role. Text and images typically convey the subjective perspective of the publisher, while comments often reflect the objective viewpoints of reviewers or readers. Existing methods, however, tend to process and fuse all three modalities indiscriminately, overlooking the nuanced relationships between subjective and objective perspectives.
This lack of differentiation not only prevents the model from effectively distinguishing between the two types of information but also introduces noise through direct fusion, which hinders judgment and degrades performance.
%In a post containing text, images, and comments, each modality serves a distinct purpose.
%Text and images reflect the subjective views of the publisher, while comments provide the more objective perspective of reviewer. The existing methods process and fuse the three modality indiscriminately, which often neglects the potential relationships between these two perspectives, making the model unable to distinguish the information from subjective and objective perspectives, and direct fusion may even interfere with the model's judgment.

To address these challenges, we propose a novel \emph{hierarchical soft prompt} model, \textsf{TriSPrompt}, consisting of three types of soft prompts: \emph{modality-aware (MA)} prompt, \emph{modality-missing (MM)} prompt, and \emph{mutual-views (MV)} prompt, that effectively detects rumors in incomplete multimodal data.
Specifically, \textit{for addressing} \textbf{CH1}, in cases of missing modalities, the MA prompt learns both heterogeneous information from specific modalities and homogeneous features from available modality data, enabling the recovery of missing modalities.
Moreover, the MM prompt integrates the missing states in incomplete multimodal data, enabling the model to better understand and adapt to the nature of modality missing information.
\textit{For addressing} \textbf{CH2}, the MV prompt reveals the potential relationships between the subjective perspectives (\textit{i.e.}, text and image) provided by the publisher and the objective perspectives (\textit{i.e.}, comment) contributed by other reviewers in the post, to effectively detect rumors.
%The ingenious application of these three prompts enables our method to simultaneously focus on incomplete modalities and the nature of rumors.
%Experimental results and visualizations show that our method performs well on three real-world benchmark datasets.
% \par (1) We take into account the general modality missing situations, which is more in line with real-world problems.
% \par (2) We propose a novel hierarchical soft prompt method, \textsf{TriSPrompt}, which considers both the incomplete modalities and the nature of rumors simultaneously.
% \par (3) Extensive experiments on there real-world benchmarks demonstrate the effectiveness of \textsf{TriSPrompt} in distinguishing rumors compared with other methods.
The main contributions of this paper can be summarized as follows:
\begin{itemize}
% \item {\color{blue}{We propose a novel \emph{hierarchical soft prompt} model, TriSPrompt, designed to effectively detect rumors in incomplete multimodal data.}}
% \item We propose a novel \emph{hierarchical soft prompt} model, \textsf{TriSPrompt}, designed to effectively detect rumors in incomplete multimodal data. To the best of our knowledge, this is the first approach to tackle the general modality-missing problem with incomplete data in both training and testing phases.
\item We propose a novel \emph{hierarchical soft prompt} model, \textsf{TriSPrompt}, designed to effectively detect rumors in incomplete multimodal data.
To the best of our knowledge, this is the first approach to tackle the general modality-missing problem with incomplete data in both training and testing phases in the field of multimodal rumor detection.
% \item  \textcolor{red}{ Specifically, the MA prompt captures heterogeneous and homogeneous features from available modalities to recover missing data, while the MM prompt models missing states to enhance adaptability to incomplete information. The MV prompt bridges subjective and objective perspectives to improve rumor detection.}

%zero-training and interpretable EC system, \textsf{ZeroEC}, designed to efficiently and effectively correct erroneous values.
%We address general scenarios involving missing modalities in both training and testing.
%To the best of our knowledge, this is the first multimodal rumor detection method that effectively handles incomplete modality in both stages.
% \item {\color{red}We introduce a novel hierarchical soft prompt model, \textsf{TriSPrompt}, which simultaneously accounts for both incomplete modalities and the dual subjective and objective perspectives nature of rumors.}
\item  In \textsf{TriSPrompt}, the MA prompt extracts features from available modalities to restore missing ones, while the MM prompt models missing states to enhance adaptability to incomplete information, effectively addressing the issue of information loss caused by modality absence, thus solving the problem of information loss caused by incomplete modalities.
\item The MV prompt of \textsf{TriSPrompt} uncovers latent relationships between the subjective perspectives (\textit{i.e.}, text and image) provided by the publisher and the objective perspectives (\textit{i.e.}, comments) from other reviewers, facilitating efficient rumor detection.
\item Extensive experiments on three real-world benchmark datasets demonstrate the effectiveness of \textsf{TriSPrompt} in distinguishing rumors, outperforming existing methods.
\end{itemize}

\section{Related Work}
\subsection{Unimodal Rumor Detection}
Research on unimodal rumor detection has primarily focused on image-based ones and text-based ones.
Image-based rumor detection methods \cite{qi2019exploiting,jin2016novel,li2015survey,gupta2013faking} focus on evaluating the logical consistency of image content and detecting signs of modifications, such as artificial splicing. These methods aim to detect whether image content is logically coherent and free from manipulations that suggest tampering or forgery.
Text-based methods \cite{ma2019detect,de2018attending,chen2018call} focus on detecting rumors by analyzing linguistic features, emotions, and writing styles \cite{rubin2015towards,przybyla2020capturing}, aiming to classify content as rumor or non-rumor. With the evolution of social media, metadata has become an important feature in rumor detection. Metadata, such as the identity of the poster, dissemination pathways, and engagement metrics like likes and shares, has significantly enhanced the ability of models to detect rumors \cite{ma2017detect,lao2021rumor}. This approach exploits the wealth of contextual data available on social platforms to enhance the performance of rumor detection models. 
% Recently, there has been a growing interest in the field of multi-domain rumor detection. Previous methods \cite{ma2016detecting} based on single-domain detection often perform poorly when generalized across domains, such as from health-related rumors to terrorism-related rumors. This occurs due to the disparate data distributions across domains, a challenge known as domain shift. Existing techniques for detecting rumors in multi-domain scenarios often struggle with severe domain transfer challenges.
\citet{nan2021mdfend} leverages a mixture-of-experts approach to extract modality-specific representations and integrates them using domain-specific gates for identifying fake news. In parallel, \citet{zhu2022memory} introduces a domain adapter combined with a memory bank to effectively address the challenges posed by domain shifts and incomplete domain labeling, enhancing the robustness of fake news detection models across varying contexts.

\subsection{Multimodal Rumor Detection}
Multimodal information dissemination has become the dominant form of communication, resulting in increased research interest in multimodal rumor detection.
Early studies often used a simplistic approach, concatenating text and image features \cite{singhal2019spotfake}, without fully leveraging the potential of multimodal data.
\citet{qi2021improving} introduces a similarity-based model for fake news detection, which computes the similarity between multimodal and cross-modal features.
In addition, \citet{wu2021multimodal} advances feature integration across modalities by employing cross-attention mechanisms, improving the fusion of multimodal data.
\citet{chen2022cross} solves the cross-modal ambiguity learning problem by quantifying the ambiguity between different unimodal features using the distribution divergence. \citet{zheng2022mfan} presents a GAT-based model that integrates text, image content, and comment graphs for rumor detection.
Recently, \citet{dhawan2024game} proposed GAME-ON, a novel end-to-end trainable GNN-based framework that allows for granular interaction modalities to fuse them early in the framework. Despite these advances, existing methods often overlook the unique properties of rumors; Text and images typically convey the subjective perspective of the publisher, while comments often reflect the objective viewpoints of reviewers or readers. Current approaches treat these three modalities as a single perspective, neglecting the intrinsic relationships between perspectives and potentially leading to misleading model judgments. 

Given the rapid spread of rumors and the common occurrence of partial modality loss during data collection, \citet{xu2023leveraging} address the issue of incomplete modalities with a novel framework that leverages contrastive learning and knowledge distillation, filling a gap in the field. However, the reliance on a teacher-student distillation structure is limited due to its dependency on complete training data and structural inflexibility. To overcome these challenges, we propose an end-to-end model with a hierarchical soft prompt architecture that simultaneously accounts for both the dual-perspectives nature of rumors and the challenges associated with incomplete training and test data.

\begin{figure*}[t]
  \includegraphics[width=\textwidth]{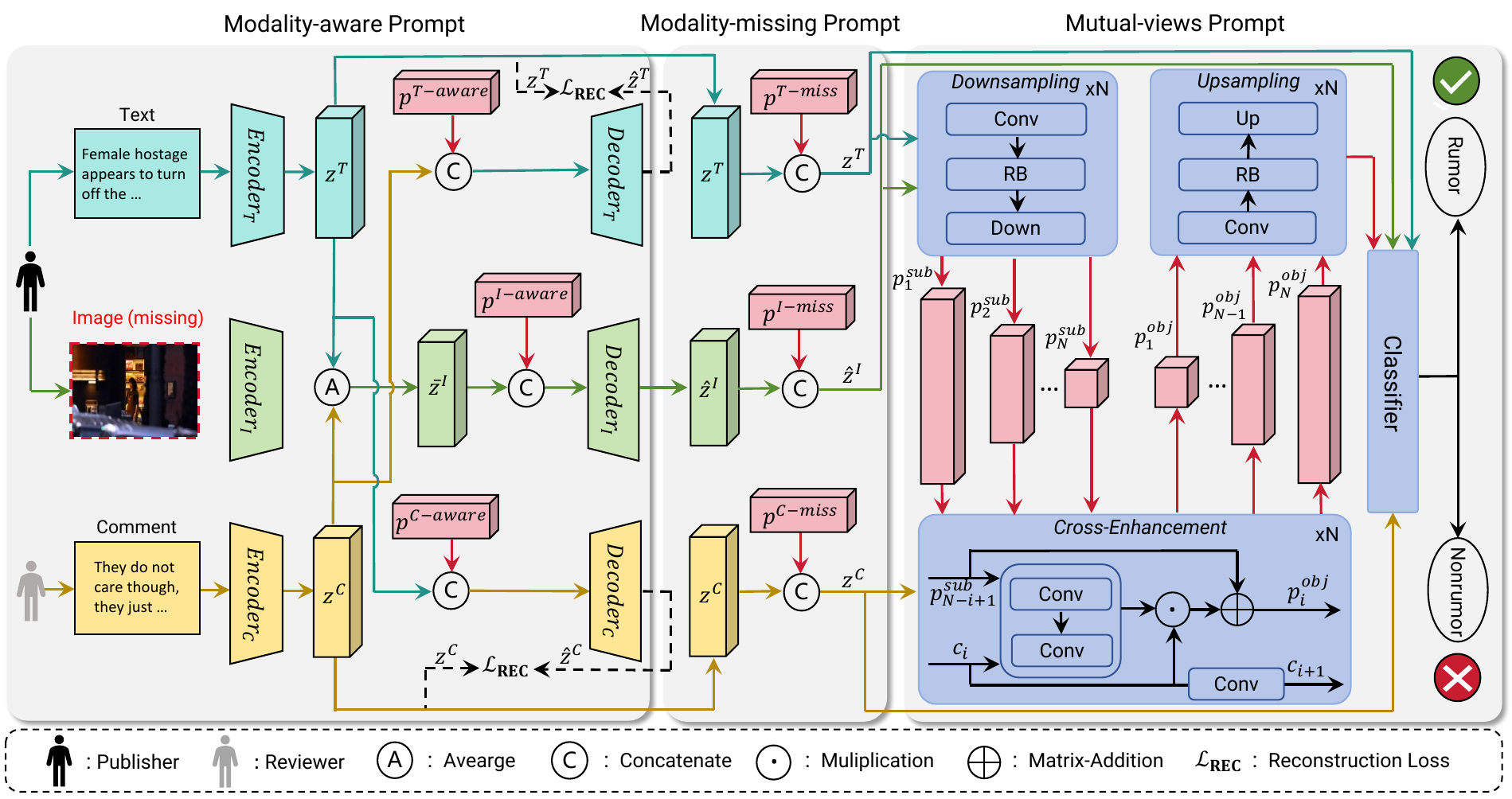}
  % \vspace*{-0.3in}
  \caption{The architecture of \textsf{TriSPrompt}. Given the input incomplete multimodal data (we assume "Image" modality is missing). It consists of modality-aware prompt, modality-missing prompt, mutual-views prompt, Conv (\emph{Convolutional-Layer}), RB (\emph{Residual-Block}), Down (\emph{Downsampling-Layer}), and Up (\emph{Upsampling-Layer}).}
  \label{fig:architecture}
  \vspace*{-0.1in}
\end{figure*}

\newtheorem{definition}{Definition}

\section{Methodology}
\subsection{Problem Formulation}
Let’s define \( P = \{p_1, \cdots ,p_n\}\) as a set of posts, where each post \( p \) consists of three modalities: \( \{T,  I , C \}\) where \(T\) donates a source text, \(I\) represents an image, and \(C\) refers to a comment.
In complete scenarios, all modalities are observed and easily integrated to support rumor detection.
However, in real-world settings, some modalities may be missing, requiring their recovery for effective fusion.
Considering the three modalities mentioned, there are seven different missing modality cases, as shown in Appendix \ref{sec:Cases}.
For simplicity, we introduce an indicator \(  \epsilon = \{ \epsilon_1,\epsilon_2,\epsilon_3\} \) for each post \( p \), where \( \epsilon_{j} = 0, j \in \{1,2,3\}\) signifies the missing of the \( j \)-th modality in the sample, and a non-zero value signifies its availability. This definition is consistent across both training and test sets.
%Our task is to detect binary rumors on posts with incomplete modalities. Specifically, our aim is to learn a function \( f_\theta \) such that \( f_\theta(p) \rightarrow \hat{y} \), where \( p \) represents a given multimodal post and \( \hat{y} \) is the outcome of the detection. \( \hat{y} = 1 \) denotes a rumor, and \( \hat{y} = 0 \) denotes a non-rumor.
\begin{definition}
{\bf Rumor Detection.}
Given a multimodal post \( p \) with a missing-modality indicator \(  \epsilon \), the goal of rumor detection is to classify posts with incomplete modalities as either rumors or non-rumors; that is, the aim is to learn a function \( f_\theta \) such that \( f_\theta(p) \rightarrow \hat{y} \), where \( p \) represents a given multimodal post, and \( \hat{y} \) is the outcome of the detection. Here, \( \hat{y} = 1 \) denotes a rumor, and \( \hat{y} = 0 \) denotes a non-rumor.
\vspace{-0.05in}
\end{definition}
\subsection{Network Overview}
% \vspace{-0.0in}
Our proposed \textsf{TriSPrompt} framework, as depicted in Figure \ref{fig:architecture}, consists of three types of soft prompts: \emph{modality-aware} (MA) prompt, \emph{modality-missing} (MM) prompt, and \emph{mutual-views} (MV) prompt.

Specifically, in \textsf{TriSPrompt}, all available modalities are first encoded into a latent space, from which features are extracted via sampling.
To account for the heterogeneity across modalities, we propose the MA prompt to learn the unique characteristics of each modality.
For missing modalities, the MA prompt employs the available modality features to reconstruct ones.
This ensures that each post is represented by a complete set $\{T, I, C\}$.
To enhance the model's awareness of missing modalities, the MM prompt indicates which modality is missing and has been reconstructed.
Subsequently, the MV prompt leverages all available modalities for comprehensive fusion, enhancing the rumor detection classification process.
Further details are provided in subsequent sections.
For simplicity, but without loss of generality, we assume the absence of the "Image" modality, \textit{i.e.}, $p = \{T, I(missing), C\}$ as illustrated in Figure \ref{fig:architecture}.
% \vspace*{-0.25in}
% \epsilon=\{1,0,1\} 

\subsection{Modality-Aware Prompt}
The interplay between modality-specific heterogeneity and homogeneity has long been a key focus in multimodal research. Ideally, models should extract homogenous features across different modalities to reduce redundancy and introduce heterogeneous features for complementarity, thus fully leveraging the advantages of multimodal data. When modalities are missing, restoring the missing modality often relies on the homogeneity of available modalities, potentially overlooking their heterogeneity. In light of these challenges, we propose the \emph{modality-aware} (MA) prompt, which is designed to capture the heterogeneous features of each modality, thereby improving the accuracy of missing modality restoration. 

Specifically, in the MA prompt, the available modalities $T$ and $C$ are first encoded into a latent space and obtain a representation $z^T$ and $z^C$, \textit{i.e.},
\begin{align}
z^T &\sim N(\mu^T, \sigma^T) = Encoder_T(T), \\
z^C &\sim N(\mu^C, \sigma^C) = Encoder_C(C), 
\end{align}
where $Encoder_T$ and $Encoder_C$ refer to a generic encoder architecture, potentially composed of multiple layers of Transformers for each modality.
$\mu^T$ and $\mu^C$ represent the mean values, while $\sigma^T$ and $\sigma^C$ represent the variances.
We adopt the \emph{variational autoencoder} (VAE) paradigm \cite{kingma2013auto}, constraining the encoding results to follow a normal distribution. This constraint leads to the \emph{Kullback–Leibler} (KL) divergence loss \(\mathcal{L}_{\textbf{KL}}\), which quantifies the gap between the latent representation and the normal distribution, \textit{i.e.},
\begin{align}\nonumber
\mathcal{L}_{\textbf{KL}} = & [\text{KL}(N(\mu^T, \sigma^T) \parallel N(0, 1)) + \notag \\\nonumber
&\quad \text{KL}(N(\mu^C, \sigma^C) \parallel N(0, 1))]/2,
\end{align}
For the missing modality $I$, we utilize the available modality features, namely $z^T$ and $z^C$, for recovery. First, we obtain the modality homogenous feature $\overline{z}^I$ by averaging.
For each modality, there exists a prompt designed to learn heterogeneous features, denoted here as $p^{x-aware} \in \mathbb{R}^{1 \times L^{aware}}$ with $x \in \{T, I, C\}$, where $L^{aware}$ donates the length of $p^{x-aware}$. We combine the homogeneous feature $\overline{z}^I$ with $p^{I-aware}$ and decode it to restore the missing modality feature $\hat{z}^I$, \textit{i.e.},
\begin{align}
~~~~~~&\overline{z}^I =[z^T+z^C]/2,\\
&\hat{z}^I=Decoder_I(\overline{z}^I,p^{I-aware}),
\end{align}
where $Decoder_I$ denotes a generic decoder architecture, similar to $Encoder$. Due to the incomplete nature of the training data, direct supervision of the reconstruction is not feasible. Thus, we reconstruct the available modalities and apply constraints to ensure consistency, \textit{i.e.}, 
\begin{align}
\hat{z}^T&=Decoder_T(z^C,p^{T-aware}),\\
\hat{z}^C&=Decoder_C(z^T,p^{C-aware}),
\end{align}

Finally, we introduce a \emph{multimodal reconstruction} loss function $\mathcal{L}_{\textbf{REC}}$ that uses \emph{mean squared error} (MSE) to quantify the distance between the original features ($z^T$ and $z^C$) and their reconstructed counterparts ($\hat{z}^T$ and $\hat{z}^C$), \textit{i.e.}, 
\begin{align}\nonumber
\mathcal{L}_{\textbf{REC}}=\left[MSE(\hat{z}^T, z^T)+MSE(\hat{z}^C, z^C)\right]/2.
\end{align}

\vspace*{-0.05in}
\subsection{Modality-Missing Prompt}
Using the MA prompt, we obtain a complete multimodal representation \( z = \{z^T, \hat{z}^I, z^C\} \), where \( z^T \) and \( z^C \) are the features from available modalities, and \( \hat{z}^I \) is the feature from the recovered modality.
To enhance the model's ability to discern between original and recovered modalities, we introduce the \emph{modality-missing} (MM) prompt as indicators. The detailed procedure is as follows:
\begin{align}
z^T&=Concat(z^T,p^{T-miss}),\\
\hat{z}^I&=Concat(\hat{z}^I,p^{I-miss}),\\
z^C&=Concat(z^C,p^{C-miss}),
\end{align}
where $p^{x-miss} \in \mathbb{R}^{1 \times L^{miss}}$ with $ x\in\{T,I,C\}$ refers to the MM prompt. $L^{miss}$ donates the length of $p^{x-miss}$. By utilizing this prompt, we enhance the representation of modality features by incorporating the missing states of incomplete multimodal data, enabling the model to more effectively recognize and adapt to modality absence.
This improves the model’s robustness in handling incomplete data.
\vspace*{-0.2in}

\subsection{Mutual-Views Prompt}
Traditional multimodal representations often overlook the distinct characteristics of rumors by applying overly simplistic or uniform fusion strategies to text, image, and comment features. A post typically comprises three components: text, image, and comment, each performing distinct functions. The text and image typically represent the publisher's subjective view, reflecting the content creator's perspective, while the comment represents the objective views of other reviewers interacting with the post. The latent relationship between these two perspectives is crucial for rumor detection.

To this end, we propose the \emph{Mutual-Views} (MV) prompt to uncover latent relationships between dual perspectives in rumor propagation, where the text and image represent the publisher's subjective view of the post, and comments reflect the objective views of other reviewers.

\emph{Architecture}. The MV prompt mainly consists of three modules, \textit{i.e.}, a \emph{downsampling} module, a \emph{cross-enhancement} module, and a \emph{upsampling} module.
Specifically, the \emph{downsampling} module integrates the subjective view information from both the text and the image.
First, it employs a \emph{cross-attention} mechanism \cite{lu2019vilbert} to fuse these modalities, capturing the interdependencies between textual and visual features. This integration is formulated as:
\begin{align}
p_1^{\text{sub}} = \text{CrossAttn}\left(z^T, \hat{z}^I\right),
\end{align}
where $z^T$ and $\hat{z}^I$ represent the text embedding and image embedding, respectively.
Subsequently, each network layer of the $N$-layer \emph{downsampling} module comprises a convolutional layer, a residual block, and a downsampling layer, which refine features while progressively reducing dimensionality.
For the $i$-th network layer $F^{down}_{i}$, the specific calculation process is  $p_{i+1}^{sub}=F^{down}_{i}(p_{i}^{sub})$, where $i={1, \cdots, N}$.
% \begin{align}\nonumber
%     p_{i+1}^{sub}=F^{down}_{i}(p_{i}^{sub}),  ~~~~i={1, \cdots, N},
% \end{align}
$p_{i}^{sub}$ denotes the output of the $i$-th layer. 
This iterative process effectively distills the fused subjective-view information, enhancing the representation for subsequent analysis.

To generate the \emph{objective-view} prompt, we use an $N$-layer \emph{cross-enhancement} module that hierarchically integrates the subjective-view prompt with comment embedding, with each layer computed as:
\begin{align}
    w_i&= Conv(p_{N-i+1}^{sub},c_i),\\
     p_{i}^{obj} &= p_{N-i+1}^{sub}+c_i \cdot w_i,\\
     c_{i+1}&= Conv(c_i),
\end{align}
where $p_{i}^{obj}$ denotes the $i$-th enhanced objective-view prompt. $Conv$ denotes the convolution operation. $c_i$ denotes the comment embedding.
By retaining intermediate results at each layer, we obtain a sequence of enhanced objective-view prompts, \textit{i.e.}, \( p^{obj} = \{p_1^{obj}, \cdots, p_N^{obj}\} \), where \(c_{1} = z^C\).
After generating the subjective-view and objective-view prompts, we perform fine-grained fusion to integrate these perspectives, enabling comprehensive modeling of rumor characteristics.

The fused representations are subsequently processed through $N$ \emph{upsampling} modules.
Each network layer comprises a convolutional layer, a residual block, and an upsampling layer.
For the $i$-th network layer $F^{up}_{i}$, 
\begin{equation}\nonumber
\begin{split}
        p_{i+1}^{mutual}=&F^{up}_{i}(p_{i}^{obj}, p_{i}^{mutual}), \quad
        i={1, \cdots, N},
\end{split}
\end{equation}
initialized with $p_{1}^{mutual}=p_{N}^{sub}$.
This refinement cascade produces the integrated prompt: $p^{mutual} \in \mathbb{R}^{1 \times L^{mutual}}$ = \( p_N^{mutual} \), where $L^{mutual}$ denotes the dimensionality of the final prompt.
The classification head processes the concatenated features: \([ p^{mutual}, z^T, \hat{z}^I, z^C]\) through fully-connected layers to classify the post as either a rumor $(\hat{y}=1)$ or non-rumor $(\hat{y}=0)$.

%Ultimately, we obtain an integrated prompt $p^{mutual} \in \mathbb{R}^{1 \times L^{mutual}}$ = \( p_N^{mutual} \) that deeply learns the underlying relationships between subjective and objective perspectives.
%$L^{mutual}$ donates the length of $p^{mutual}$.
%After the fusion, the integrated prompt \( p^{mutual} \), along with the text embedding \( z^T \), image embedding \( \hat{z}^I \), and comment embedding \( z^C \), are passed through the classification head for rumor detection.
%The model classifies the post as a rumor $(\hat{y}=1)$ or non-rumor $(\hat{y}=0)$.

\emph{Loss Function}. 
In general, we introduce a rumor detection loss function $\mathcal{L}_{\textbf{CLS}}$ that uses \emph{cross-entropy} (CE) loss for binary classification to train the model, which helps optimize the parameters by minimizing the difference between the predicted $\hat{y}$ and the true label $y$, \textit{\textit{i.e.}},
\begin{align}\nonumber
&\mathcal{L}_{\textbf{CLS}} = CE(\hat{y}, y),
%\\\nonumber
%&= -\frac{1}{N} \sum_{i=1}^{N} \left[ y_i \log(\hat{y}_i) + (1 - y_i) \log(1 - \hat{y}_i) \right]\nonumber
\end{align}
where $\hat{y} = Classifier(z^T, \hat{z}^I, z^C, p^{mutual})$.
%Ground truth label for the ii-th sample
The $Classifier$ is a multimodal rumor detection classifier composed of multi-layer perceptions.

\subsection{Objective Function}
%\emph{Overall objective}.
The overall objective function $\mathcal{L}_{\textbf{Total}}$ of our model \textsf{TriSPrompt} contains three types of losses, including the rumor detection loss $ \mathcal{L}_{\textbf{CLS}}$, the KL divergence loss $\mathcal{L}_{\textbf{KL}}$ and the multimodal reconstruction loss $\mathcal{L}_{\textbf{REC}}$, \textit{\textit{i.e.}},
\begin{equation}\nonumber
    \mathcal{L}_{\textbf{Total}} = \mathcal{L}_{\textbf{CLS}}+ \lambda_1 \cdot \mathcal{L}_{\textbf{KL}} + \lambda_2 \cdot \mathcal{L}_{\textbf{REC}},
\end{equation}
where \(\lambda_1\) and \(\lambda_2\) are trade-off hyperparameters.

\section{Experiments}
\label{sec:EXperiments}
% In this section, we evaluate the performance of our proposed model \textsf{TriSPrompt} and five state-of-the-art methods.
% All methods were implemented in Python.
% The experiments were conducted in an Intel Core 2.90GHz server with A40 48GB (GPU) and 256GB RAM, running Ubuntu 18.04 system.

%\subsection{Datasets and Implementation Details}
~~ \textbf{Datasets.} In our experiments, we utilize three multimodal rumor detection datasets: the Chinese datasets \emph{Weibo-19} \cite{song2019ced} and \emph{Weibo-17} \cite{jin2017novel}, along with the English dataset \emph{Pheme} \cite{zubiaga2017exploiting}.
Details are in Appendix \ref{sec:appendix}.
Each dataset comprises source text and images; however, \emph{Weibo-19} and \emph{Pheme} include comments, whereas \emph{Weibo-17} does not.
For \textsf{TriSPrompt}, we directly use \(p_1^{sub}\) obtained from text and image to replace the MV prompt on \emph{Weibo-17} dataset.
%Table \ref{tab:dataset} lists the information of each dataset, including the number of non-rumors (\#Ns), rumors (\#Rs), samples (\#Tuples), respectively.

\begin{table*}[]
    \centering
        %MRD (Multimodal Rumor Detection) and IML (Incomplete Modality Learning).}
        %Bold is the best.}
    \setlength{\tabcolsep}{2pt}
    \fontsize{8pt}{13pt}\selectfont
    \begin{tabular}{c|c|ccc|ccc|ccc}
    \hline
    \multicolumn{2}{c}{\multirow{2}{*}{\textbf{Methods}}}&\multicolumn{3}{|c}{\emph{Weibo-19}}&\multicolumn{3}{|c}{\emph{Pheme}}&\multicolumn{3}{|c}{\emph{Weibo-17}} \\
    \cline{3-11}
     \multicolumn{2}{c|}{}&ACC&F1&AUC&ACC&F1&AUC&ACC&F1&AUC\\
    \hline
     \hline
      \multirow{3}{*}{MRD}&CAFE&67.89$\pm$0.83&44.89$\pm$2.65&0.673$\pm$0.018&61.65$\pm$0.14&60.16$\pm$3.15&0.599$\pm$0.022&65.02$\pm$1.39&65.51$\pm$5.02&0.622$\pm$0.036 \\
     &CLKD&69.19$\pm$1.64&30.68$\pm$4.85&0.574$\pm$0.048&64.18$\pm$0.84&67.31$\pm$1.63&0.616$\pm$0.016&63.26$\pm$0.70&62.10$\pm$1.55&0.574$\pm$0.028 \\
     &GameOn&76.07$\pm$0.62&56.98$\pm$1.99&0.747$\pm$0.062&66.22$\pm$0.43&71.65$\pm$0.49&0.702$\pm$0.011&66.78$\pm$1.44&64.81$\pm$6.38&0.739$\pm$0.008 \\
    \hline
     \multirow{2}{*}{IML}&Dicmor&77.48$\pm$0.85&68.73$\pm$1.88&0.823$\pm$0.021&73.17$\pm$0.72&76.10$\pm$0.97&0.776$\pm$0.008&72.58$\pm$0.80&72.48$\pm$2.41&0.789$\pm$0.006 \\
     &RedCore&80.05$\pm$0.73&70.82$\pm$2.45&0.846$\pm$0.026&71.58$\pm$0.12&75.65$\pm$0.40&0.745$\pm$0.002&73.68$\pm$0.57&73.62$\pm$0.13&0.816$\pm$0.011 \\
     \hline
     \multirow{2}{*}
        {MLLM}&Qwen-VL&61.23$\pm$0.67&38.46$\pm$2.12&0.597$\pm$0.011&50.83$\pm$0.25&23.59$\pm$0.69&0.516$\pm$0.002&50.47$\pm$0.23&21.48$\pm$2.88&0.504$\pm$0.017 \\
     &GPT-4o&64.61$\pm$0.11&41.13$\pm$1.54&0.642$\pm$0.001&47.98$\pm$0.36&22.29$\pm$0.59&0.501$\pm$0.004&56.47$\pm$0.13&24.21$\pm$1.12&0.556$\pm$0.007 \\
    \hline
      \multicolumn{2}{c|}{\textsf{TriSPrompt}}&\textbf{81.51$\pm$0.59}&\textbf{73.38$\pm$0.81}&\textbf{0.867$\pm$0.023}&\textbf{75.82$\pm$0.44}&\textbf{77.33$\pm$0.69}&\textbf{0.788$\pm$0.015}&\textbf{75.15$\pm$0.24}&\textbf{74.17$\pm$1.67}&\textbf{0.836$\pm$0.016} \\
    \hline
    \end{tabular}
            \caption{Evaluation results on three datasets at $R_m$ = 0.5, including three metrics: ACC, F1, and AUC.}
        % \vspace*{-0.05in}
    % \vspace*{-0.1in}
    \label{tab:results}
\end{table*}

 \begin{figure*}[t]
  \includegraphics[width=0.98\textwidth]{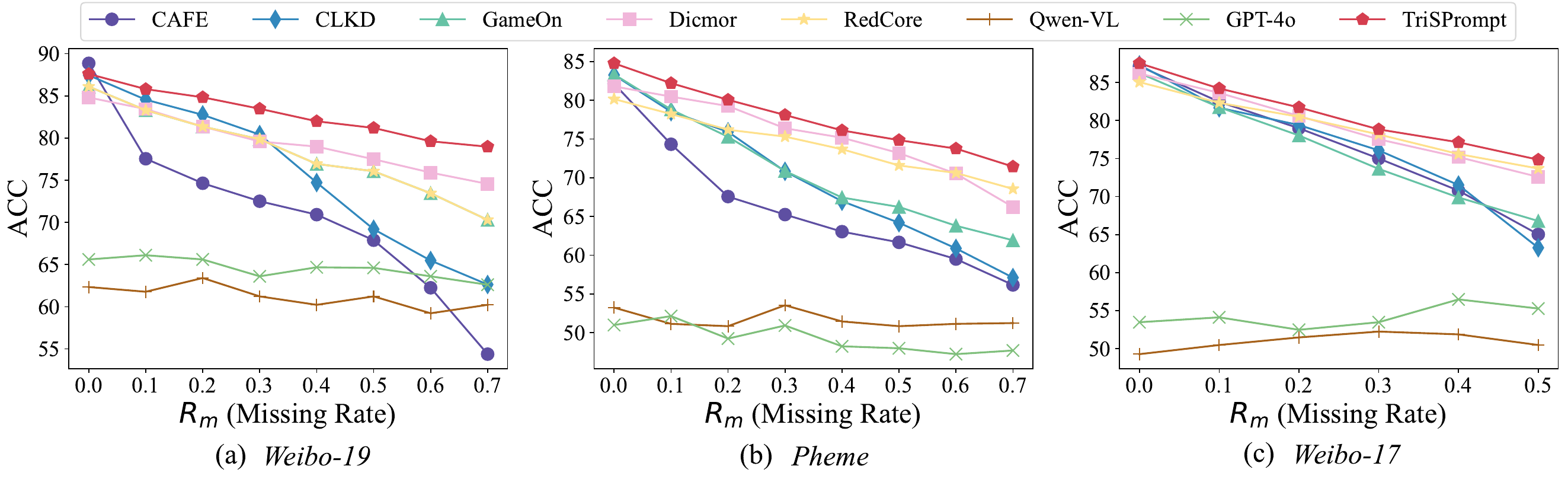}
  % \vspace*{-0.1in}
  \caption{Performance demonstration under different missing rates.}
 \vspace*{-0.15in}
  \label{fig:mr}
\end{figure*}
\textbf{Metrics.}
To evaluate and compare our framework with others, we employ standard metrics for binary classification, including Accuracy, F1 score, and AUC.
For feature extraction, we utilize a pre-trained BERT model \cite{kenton2019bert} to process text and comment information, generating a 768-dimensional hidden state as the feature representation.
Similarly, for image data, we use a pre-trained ResNet-34 model \cite{he2016deep}, which encodes the visual information into a 512-dimensional hidden state representation.

\textbf{Baselines.}
We compare \textsf{TriSPrompt} with three state-of-the-art \emph{Multimodal Rumor Detection} (MRD) approaches: CAFE \cite{chen2022cross}, CLKD \cite{xu2023leveraging}, and Game-On \cite{dhawan2024game}; two \emph{Incomplete Modality Learning} (IML) methods: Dicmor \cite{wang2023distribution} and RedCore \cite{sun2024redcore}; and two \emph{Multimodal Large Language Models} (MLLMs): Qwen2.5-VL-72B \cite{bai2025qwen2} and GPT-4o \cite{hurst2024gpt}.

\textbf{Implementation details.}  To simulate real-world modality dropout, we implement a random missing protocol \cite{wang2024incomplete}, where each complete post is subject to the random absence of one or two modalities.
This dropout mimics the incomplete information often present in real-world social media posts, making the model more robust to missing data. We denote the global missing rate as $R_m$  to estimate the prevalence of such absences. The $R_m$ is defined as $R_m = 1 - \frac{1}{L \times M} \sum_{i=1}^{L} a_i$, 
% \begin{align}
%     R_m = 1 - \frac{1}{L \times M} \sum_{i=1}^{L} a_i
% \end{align}
where \(a_i\) denotes the number of available modalities for the \(i\)-th sample, \(L\) denotes the total number of samples, and \(M\) indicates the number of modalities.
We also ensure that at least one modality is available for each sample, so \(a_i \geq 1\) and \(R_m \leq \frac{M-1}{M}\). 
The details are shown in Appendix \ref{sec:RMP}.
For three modalities $\{T, I, C\}$ datasets \emph{Weibo-19} and \emph{Pheme} (text, image, and comment), we choose the $R_m$ from \([0.0, \cdots, 0.7]\), where 0.7 is an approximation of \(\frac{M-1}{M}\) with the same meaning.
For two modalities $\{T, V\}$ dataset \emph{Weibo-17} (text, image), which only involves the scenario of one modality being missing, we choose the $R_m$ from \([0.0, \ldots, 0.5]\).
% For \textsf{TriSPrompt}, we directly use \(p_1^{sub}\) derived from the text and image, to replace the MV prompt.
We use Adam optimizer with ${\beta}_1$ = 0.99 and ${\beta}_2$ = 0.999. The learning rate is set to 0.002.
The training batch size is set to 64.
Number of network layers is set to 3.
The hyperparameters \(\lambda_1 = 0.1\) and \(\lambda_2 = 0.001\). For fairness, all results are averaged over five times.
\vspace*{-0.25in}
\begin{figure*}[t]
  \includegraphics[width=0.97\textwidth]{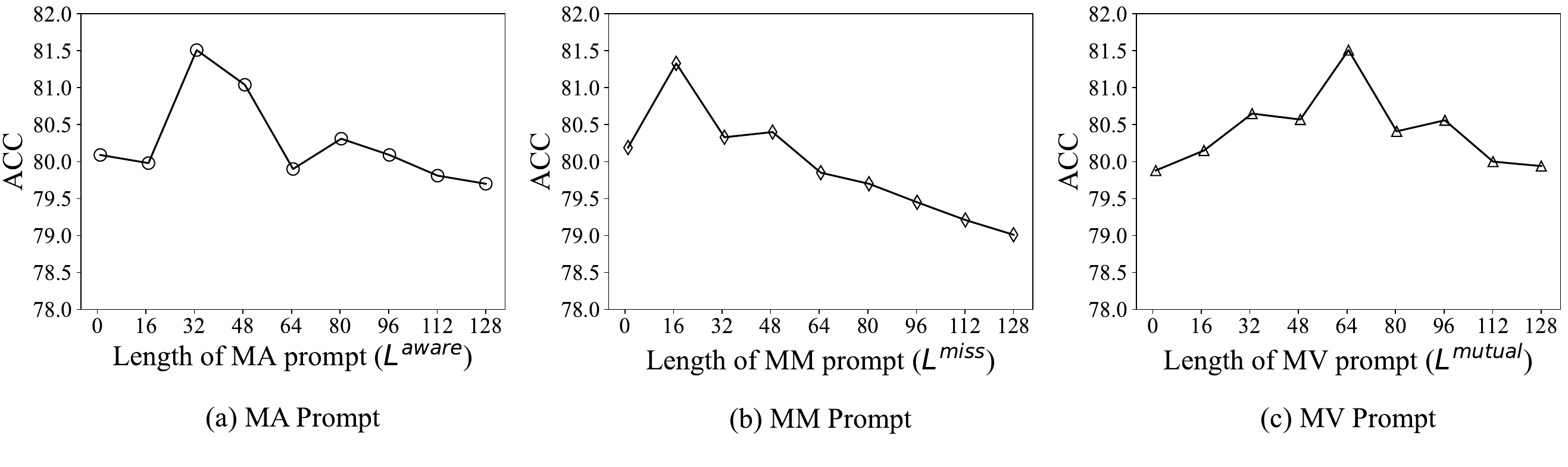}
  % \vspace*{-0.15in}
  \caption{Ablation study on different lengths of prompts on the \emph{Weibo-19} dataset.}
  \label{fig:len}
   \vspace*{-0.15in}
\end{figure*}

\subsection{Comparison Study}
Table \ref{tab:results} presents the accuracy of various rumor detection methods on three publicly real-world datasets under a global missing rate of $R_m$ = 0.5, indicating that half of the modalities are missing.

One can observe that, \textsf{TriSPrompt} significantly outperforms all of the baselines.
In terms of prediction accuracy (\textit{i.e.}, ACC, F1, and AUC), \textsf{TriSPrompt} exceeds the best-performing baseline, RedCore, by an average of 2.99\%, and even increases up to 5.92\% on the \emph{Pheme} dataset w.r.t. Accuracy.
In particular, compared with MRD based methods, \textsf{TriSPrompt} demonstrates significantly superior performance.
It outperforms the best MRD based method, GameOn, by an average of 14.08\%, and even increases up to 28.78\% on the \emph{Weibo-19} dataset w.r.t. F1 score.
This is because \textsf{TriSPrompt} fully leverages the available modalities, integrating heterogeneous and homogeneous features to effectively reconstruct missing data. We provide detailed experimental verification in Appendix \ref{sec:RMM}. to support this claim.
It also perceives and adapts to the modality-missing mechanism while deeply exploring both subjective and objective perspectives in rumors. Although MLLMs exhibit strong capabilities in cross-modal understanding, their performance on rumor detection remains suboptimal compared to traditional methods. This primarily stems from the fact that MLLMs commonly treat input as an intrinsic representation of "reality", without incorporating dedicated mechanisms to evaluate content veracity—a crucial component for robust rumor detection.
% \vspace*{-0.15in}

% This is primarily because MLLMs are generally tend to assume the input information is inherently "Reality", lacking an explicit mechanism for assessing the veracity of the content — a critical capability for effective rumor detection.}

% At MR=0.5, half of the modalities in all samples are missing, a severe yet inevitable scenario in reality. Observing the results in Table 1, our method maintains leading performance across all evaluation metrics on three datasets. For example, state-of-the-art methods like Game-On experience a significant accuracy drop of over 10\% when half of the modalities are missing, highlighting their struggle to adapt to modality absence. In contrast, \textsf{TriSPrompt} maintains strong performance, with only a slight decrease in accuracy, demonstrating its robustness to missing modalities. This phenomenon suggests that these models fail to adapt to such absences and cannot effectively utilize the available modality information for rumor detection classification, whereas our method can. Furthermore, compared to general Incomplete Modality Learning methods, our performance is also superior, as we have tailored our design according to the nature of rumor detection tasks. In a sense, our method is the optimal rumor detection approach under incomplete modalities.

\subsection{Parameter Evaluation}
\vspace*{-0.01in}
\emph{Effect of missing rate.}
The experimental results for varying the missing rate $R_m$ from 0\% to 70\% are shown in Figure~\ref{fig:mr}. We can find that, with the growth of the missing rate, the prediction accuracy (\textit{i.e.}, ACC) of each algorithm descends consistently.
It is attributed to the less data information for detection when the missing rate turns high.
Among all the algorithms, \textsf{TriSPrompt} performs the best in each case.
% Moreover, its accuracy becomes more stable with the increase in missing rate.
% In other words, \textsf{TriSPrompt} is more robust with the increasing missing rate $R_m$ than others.
Notably, across different missing rates, the average performance degradation ratio of \textsf{TriSPrompt} is just 6.14\%, far lower than the best-performing baseline RedCore's 8.32\%.
In other words, \textsf{TriSPrompt} is more robust with the increasing missing rate than others.
The reason behind this is that, \textsf{TriSPrompt} learns the data distribution from all observed multimodal data conditional on the feature missing state, so as to decrease or delay the impact of increased missing rate on the forecasting performance.
In real-world rumor outbreaks, severe modality absence often renders existing detection algorithms ineffective, making our \textsf{TriSPrompt} approach crucial for multimodal rumor detection. Despite their flexibility in handling arbitrary modality inputs, including unimodal and multimodal content, MLLMs still fall short in achieving strong performance on rumor detection.

\emph{Effect of prompt length.}
With the prompt lengths (\textit{i.e.}, $L^{aware}$, $L^{miss}$, $L^{mutual}$) corresponding to the MA prompt, MM prompt, and MV prompt, varying from 1 to 128, Figure \ref{fig:len} illustrates the detection accuracy (\textit{i.e.}, ACC) of \textsf{TriSPrompt} on the \emph{Weibo-19} dataset.
% Experimental results indicate that the performance improves as the prompt length increases. However, after a certain length is reached, performance begins to decline, suggesting the need for a balanced length. 
The experimental results demonstrate that model performance generally improves with increased prompt length, up to a certain threshold. Beyond this point, further extension leads to a decline in accuracy, indicating that excessively long prompt may introduces noise or dilute the importance of input features. This suggests the necessity of selecting an appropriate length.
Specifically, \textsf{TriSPrompt} is the best, in terms of the larger Accuracy, when $L^{aware}$ is 32, $L^{miss}$ is 16,  and $L^{mutual}$ is 64.
It also confirms that, the three prompts of \textsf{TriSPrompt} benefit the detection accuracy.

% In addition to validating and analyzing the effectiveness of the three soft prompts, we also investigated the impact of prompt length (\textit{i.e.}, $L^{aware}$, $L^{miss}$, $L^{mutual}$) on performance. As shown in Figure \ref{fig:len}, experimental results indicate that the model's performance improves with increasing prompt length. However, when the length reaches a certain value, the model performance begins to decline. For example, when the length of MA prompt $L^{aware}$ = 32, the model performance reaches its peak. This can be attributed to the role of the prompt in enhancing the model's features. As the prompt length increases, its ability to capture and enhance relevant features improves. However, when the length of prompt exceeds a certain threshold, it may overshadow the original features, hindering the model's ability to effectively learn the data's inherent characteristics, which leads to performance degradation.

\begin{table}[t]
\centering
% \vspace*{-0.12in}
\setlength{\tabcolsep}{2pt}
\fontsize{9pt}{13pt}\selectfont
\begin{tabular}{c  | c c c c c }
\hline
\textbf{Datasets} & no-MA & no-MM & no-MV & no-MAM & \textsf{TriSPrompt} \\
\hline
\hline
% $p^{aware}$ & & $\checkmark$& $\checkmark$& &$\checkmark$\\
% $p^{miss}$ & $\checkmark$ & & $\checkmark$& &$\checkmark$\\
% $p^{mutual}  $ & $\checkmark$ &$\checkmark$  & & $\checkmark$&$\checkmark$ \\

% \hline
% \hline
\emph{Weibo-19}  & 80.14 & 80.09 &79.88 &79.17 &\textbf{81.51} \\
\emph{Pheme}  & 73.84 & 73.17 & 72.77& 71.85&\textbf{75.82} \\
\emph{Weibo-17}  &73.67 &73.01 &72.12 &70.69 & \textbf{75.15}\\
\hline
\end{tabular}
\caption{Ablation study on the hierarchical soft prompts at $R_m$ = 0.5. Quality metrics: ACC. \textbf{Bold} indicates the best performance.}
% \vspace*{-0.2in}
\label{tab:ablation}
\end{table}

\begin{table*}[h]
\setlength{\tabcolsep}{2pt}
\fontsize{9pt}{13pt}\selectfont
\centering
\begin{tabular}{c |cccc}
\hline
\textbf{Methods} & \textbf{Params (M)} & \textbf{Training Time/Epoch (ms)} & \textbf{Inference Time/Batch (ms)} & \textbf{Max GPU Memory (GB)} \\
\hline
\hline
CAFE        & 2.95 & 1996 & 910  & 0.98 \\
CLKD        & 2.81 & 1411 & 1260 & \textbf{0.84} \\
GameOn      & \textbf{1.02} & \textbf{606}  & \textbf{271}  & 1.60 \\
Dicmor      & 3.46 & 4511 & 1833 & 3.26 \\
RedCore     & 7.72 & 1923 & 822  & 2.41 \\
Qwen-VL     & --   & --   & 1455 & --   \\
GPT-4o      & --   & --   & 1623 & --   \\
\textsf{TriSPrompt} & 2.41 & 2032 & 931  & 1.28 \\
\hline
\end{tabular}
\caption{Resource consumption and computational efficiency of different methods.}
\label{tab:Eff}
\end{table*}

\subsection{Ablation Study}

We investigate the influence of different elements of \textsf{TriSPrompt} on the rumor detection performance.
The corresponding experimental results over the three datasets, including ACC, are presented in Table \ref{tab:ablation}.
no-MA is the variant of \textsf{TriSPrompt} without the MA prompt.
no-MM is the variant of \textsf{TriSPrompt} without the MM prompt.
no-MV is the variant of \textsf{TriSPrompt} without the MV prompt.
no-MAM is the variant of \textsf{TriSPrompt} without both the MA and MM prompts.

We observe that each of the three prompts in \textsf{TriSPrompt} — the MA prompt, MM prompt, and MV prompt — positively impacts detection performance.
When the three prompts are removed respectively, the detection accuracy (measured by ACC) of \textsf{TriSPrompt} decreases on average by 2.09\%, 2.70\%, and 3.35\%. In addition, when both the MA prompt and MM prompt are removed simultaneously, the model's performance experiences the largest decline, with an average decrease of 4.68\%.
%Furthermore, the MS strategy, CTS module, CS strategy, RG strategy, and ER prompt each contribute substantially to EC performance.
These findings highlight the indispensability of all three prompts in \textsf{TriSPrompt}, demonstrating their combined significance in achieving effective performance.

\subsection{Analysis of Effectiveness and Efficiency}

To rigorously assess the effectiveness and efficiency of \textsf{TriSPrompt}, we systematically compared \textsf{TriSPrompt} with representative methods from three major domains (MRD, IML, and MLLM) on the \emph{Pheme} dataset, covering model size, training/inference efficiency, memory usage, and actual detection performance presented in Table \ref{tab:Eff}. All experimental results we obtained under the same experimental environment, averaged over multiple independent runs to ensure fairness and reliability of the comparisons. Specifically, in the MRD domain, methods such as CAFE, CLKD, and GameOn mainly focus on multimodal feature fusion but have limited adaptability to missing modalities. On the Pheme dataset, \textsf{TriSPrompt} achieves an ACC of 75.82\%, which is 9.6 percentage points higher than GameOn (66.22\%). Meanwhile, \textsf{TriSPrompt}’s inference time and resource consumption are comparable to these baselines.
In the IML domain, methods such as Dicmor and RedCore focus on scenarios with missing modalities. \textsf{TriSPrompt} outperforms RedCore by 4.24\% in ACC (75.82\% vs. 71.58\%), and its performance degrades more slowly under high missing rates (average drop of 6.14\%, compared to RedCore’s 8.32\%), demonstrating stronger practical robustness.
In the MLLM domain, models like Qwen-VL and GPT-4o possess strong cross-modal understanding but perform poorly on rumor detection tasks. In contrast, \textsf{TriSPrompt} not only achieves higher detection accuracy but also consumes significantly fewer resources (e.g., inference time is only 1/17 that of GPT-4o).

% \vspace*{-0.1in}
\section{Conclusion}
% \vspace*{-0.1in}
In this paper, we innovatively propose a hierarchical soft prompt framework, namely \textsf{TriSPrompt}, targeting the more prevalent modality absence situation in multimodal rumor detection, where both the training and testing data suffer from modality missing.
It consists of the MA prompt, MM prompt, and MV prompt.
The MA prompt extracts both heterogeneous and homogeneous features from available modalities to recover missing data, while the MM prompt models missing states to improve adaptability to incomplete information. The MV prompt connects subjective and objective perspectives, enhancing rumor detection.
The effectiveness of \textsf{TriSPrompt} has been verified on three real-world datasets. Particularly, \textsf{TriSPrompt} exhibits excellent robustness under the circumstances of severe modality absence.

\section*{Limitations}
Our proposed method \textsf{TriSPrompt} focuses on the provided real-world multimodal rumor detection datasets, enabling controlled evaluation and comparison with existing approaches.
However, it falls short in addressing the complexities and variability of real-world multimodal data.
This limitation underscores the need for more robust models capable of interacting effectively with real-world scenarios.
Beyond this work, we believe some promising future works with large language models would solve this problem.

\section*{Acknowledgments}
This work was supported in part by the “Pioneer” and “Leading Goose” R\&D Program of Zhejiang (No. 2024C01212), and the Ningbo Yongjiang Talent Programme Grant 2024A-158-G.
Yangyang Wu is the corresponding author of the work.

% This document has been adapted
% by Steven Bethard, Ryan Cotterell and Rui Yan
% from the instructions for earlier ACL and NAACL proceedings, including those for
% ACL 2019 by Douwe Kiela and Ivan Vuli\'{c},
% NAACL 2019 by Stephanie Lukin and Alla Roskovskaya,
% ACL 2018 by Shay Cohen, Kevin Gimpel, and Wei Lu,
% NAACL 2018 by Margaret Mitchell and Stephanie Lukin,
% Bib\TeX{} suggestions for (NA)ACL 2017/2018 from Jason Eisner,
% ACL 2017 by Dan Gildea and Min-Yen Kan,
% NAACL 2017 by Margaret Mitchell,
% ACL 2012 by Maggie Li and Michael White,
% ACL 2010 by Jing-Shin Chang and Philipp Koehn,
% ACL 2008 by Johanna D. Moore, Simone Teufel, James Allan, and Sadaoki Furui,
% ACL 2005 by Hwee Tou Ng and Kemal Oflazer,
% ACL 2002 by Eugene Charniak and Dekang Lin,
% and earlier ACL and EACL formats written by several people, including
% John Chen, Henry S. Thompson, and Donald Walker.
% Additional elements were taken from the formatting instructions of the \emph{International Joint Conference on Artificial Intelligence} and the \emph{Conference on Computer Vision and Pattern Recognition}.

\bibliography{acl}

\appendix

\section{Missing Modality Cases }
\label{sec:Cases}
\begin{figure}[t]
\centering
  \includegraphics[width=0.9\columnwidth]{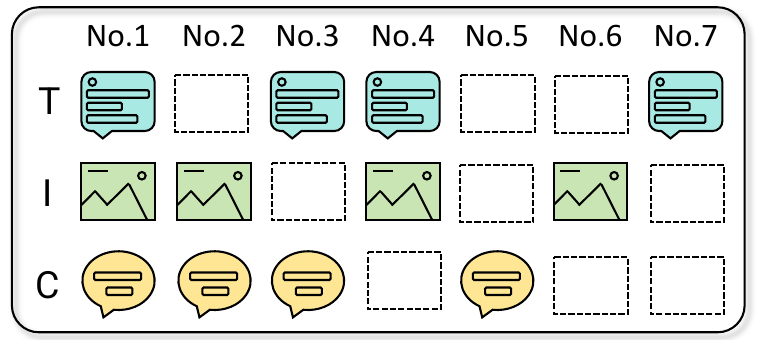}
  \caption{The seven different missing modality cases. Full modality: \{No.1\}. One modality is missing: \{No.2, No.3, and No.4\}. Two modalities are missing:\{No.5, No.6, and No.7\}. T (Text), I (Image), C (Comment).}
  \label{fig:complete}
  \vspace*{-0.1in}
\end{figure}

Considering the three modalities \( \{T,  I , C \}\), where \(T\) donates a source text, \(I\) represents an image, and \(C\) refers to a comment, as shown in Figure \ref{fig:complete}, there are seven different cases of missing modalities.
%Considering the three modalities mentioned, there are seven different missing modality cases, {\color{blue}as shown in Appendix \ref{sec:Cases}.}

\begin{figure*}[t]
\centering
  \includegraphics[width=0.75\textwidth]{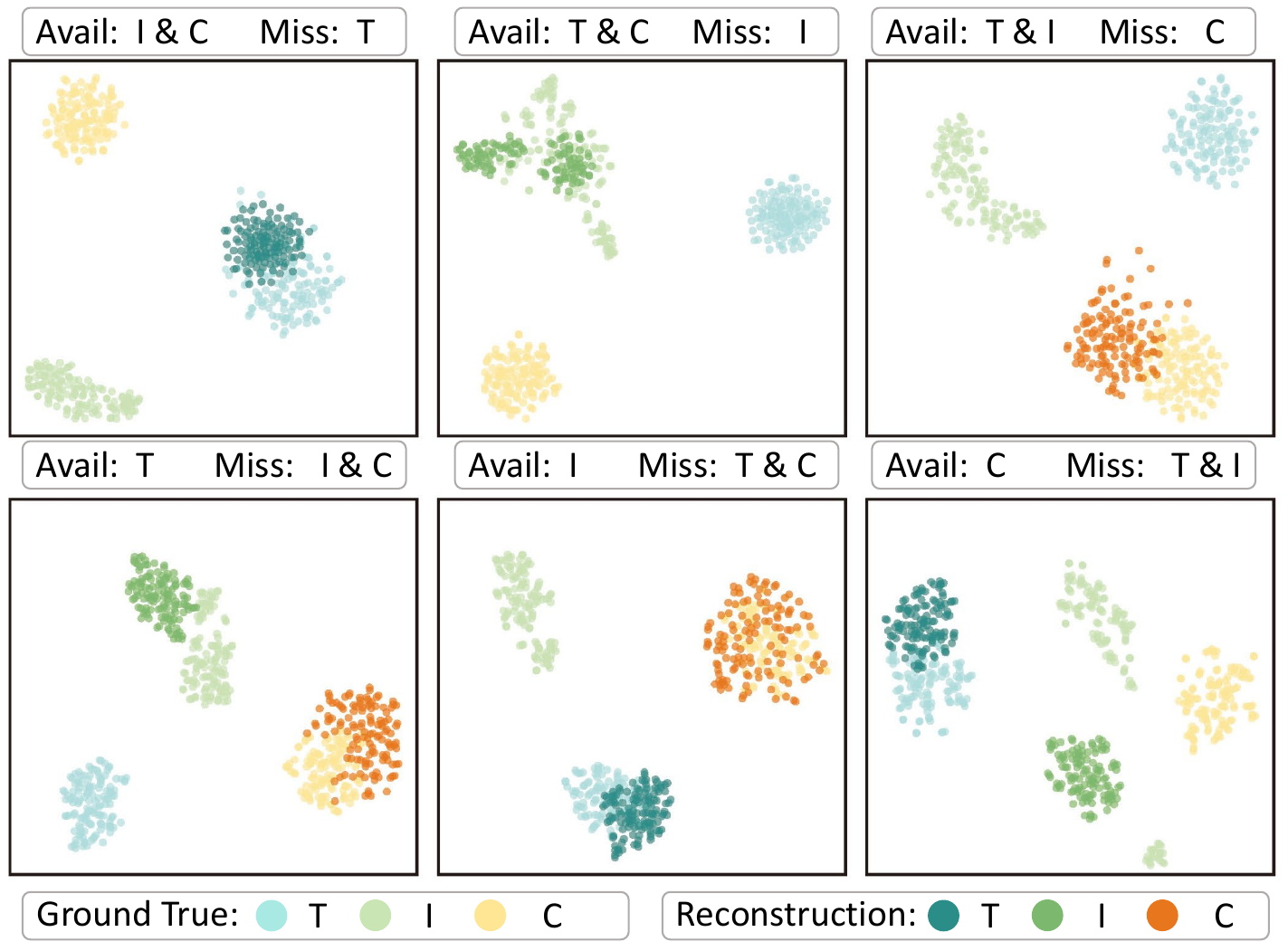}
  \caption{ Visualization of reconstructed modality feature embeddings versus ground truth under the different missing modality cases. T (Text), I (Image), C (Comment).}
  \label{fig:rmm}
  \vspace*{-0.1in}
\end{figure*}

\section{Dataset Description}
\label{sec:appendix}

\begin{table}[h]
\centering
%\#N: Non-rumors; R: Rumors.}
\setlength{\tabcolsep}{5.5pt}
\fontsize{9pt}{13pt}\selectfont
\begin{tabular}{c | c c c c}
\hline
\textbf{Datasets} & \textbf{\#Ns} & \textbf{\#Rs} & \textbf{\#Tuples}  & {\textbf{Modality types}}\\
\hline
\hline

\emph{Weibo-19}  & 877   & 590   & 1,467  & T \& I \& C  \\
\emph{Pheme}     & 1,428 & 590   & 2,018  & T \& I \& C \\
\emph{Weibo-17}  &4,749  &4,779  & 9,528  & T \& I \\
\hline
\end{tabular}
\caption{The information of each dataset or our experiments, \textit{i.e.}, the number of non-rumors (\#Ns), rumors (\#Rs), samples (\#Tuples), and modality types, respectively. T (Text), I (Image), C (Comment).}
\label{tab:dataset}
\end{table}

Table \ref{tab:dataset} lists the information of each dataset.

\begin{itemize}
    \item \textbf{\emph{Weibo-19}} dataset is collected from Weibo, one of the most popular social platforms in China. It contains 1,467 data samples, among which there are 877 non-rumor samples and 590 rumor samples.
    \item \textbf{\emph{Pheme}} dataset is constituted by tweets on the Twitter platform and based on five breaking news. It contains 2018 data samples, among which there are 1428 non-rumor samples and 590 rumor samples.
    \item \textbf{\emph{Weibo-17}} dataset is collected from the Chinese social media platform, Weibo, containing 4,749 real, 4,779 fake tweets, and 9,528 images. The fake news in the dataset was verified by the debunking system from May 2012 to January 2016.
\end{itemize}

\section{Random Missing Protocol}
\label{sec:RMP}

\begin{table}[h]
\centering
\setlength{\tabcolsep}{10pt}
\fontsize{9pt}{13pt}\selectfont
 \begin{tabular}{c|cc|cc|c}
    \hline
    \multirow{2}{*}{\textbf{$R_m$}}&\multicolumn{2}{c|}{\emph{Weibo-19}}&\multicolumn{2}{c|}{\emph{Pheme}}&\emph{Weibo-17} \\
     &$R_o$&$R_t$&$R_o$&$R_t$&$R_o$\\
    \hline
    \hline
    0.0&0.0& 0.0& 0.0& 0.0& 0.0\\
    0.1&0.1& 0.1& 0.1& 0.1& 0.2\\
    0.2&0.2& 0.2& 0.2& 0.2& 0.4\\
    0.3&0.3& 0.3& 0.3& 0.3& 0.6\\
    0.4&0.2& 0.5& 0.2& 0.5& 0.8\\
    0.5&0.1& 0.7& 0.1& 0.7& 1.0\\
    0.6&0.0& 0.9& 0.0& 0.9& $-$\\
    0.7&0.0& 1.0& 0.0& 1.0& $-$\\
   \hline

\end{tabular}
\caption{Modality-missing proportion at different missing rates. \textit{i.e.}, the global missing rate ($R_m$), one modality missing rate (c), two modality missing rate ($R_t$).}
\label{tab:RMP}
\end{table}

\begin{table}[h]
\centering
\setlength{\tabcolsep}{12pt}
\fontsize{9pt}{13pt}\selectfont 
\begin{tabular}{c |c c }
\hline 
\textbf{Case (A, M)} & \textbf{Reconstructed} & \textbf{Random} \\
\midrule
\hline 
\hline 
I \& C , T      & 0.5011 & 0.0038  \\
T \& C , I      & 0.5712 & -0.0020 \\
T \& I , C      & 0.4638 & -0.0095 \\
T , I \& C      & 0.4233 & 0.0051  \\
I , T \& C      & 0.4688 & -0.0182 \\
C , T \& I      & 0.2765 & -0.0056 \\
 \hline 
\bottomrule
\end{tabular}
\caption{
Cosine similarity results. 
Case (A, M): Available modalities (A), missing modality (M). 
Recon: Cosine similarity (Reconstructed vs GT). 
Random: Cosine similarity (Random vs GT).
 T (Text), I (Image), C (Comment).}
\label{tab:COS}
\end{table}

In the experiments, we adopted the Random Missing Protocol mechanism to simulate the modality-missing situations in the real world, where each complete post is subject to the random absence of one or two modalities. For the three-modality datasets \emph{Weibo19} and \emph{Pheme}, we choose the $R_m$ from \([0.0, \cdots, 0.7]\). For the two-modality dataset \emph{Weibo17}, there are only samples with one modality missing and we choose the $R_m$ from \([0.0, \cdots, 0.5]\). Table \ref{tab:RMP} shows the corresponding relationships between the proportion of samples with one or two modalities missing and the total number of samples at specific missing rates.

\section{Reconstruct Missing Modality}
\label{sec:RMM}

Figure \ref{fig:rmm} visualizes the distribution of reconstructed modality feature embeddings versus ground truth under different missing modality cases. We randomly sample 128 instances in the testing set from the \emph{Pheme} dataset for this comparison. The features of the selected samples are projected into a 2D space by t-SNE \cite{van2008visualizing}. From visualization results, we observe that under varying modality-missing scenarios, \textsf{TriSPrompt} successfully reconstructs missing modality feature embeddings by synergistically leveraging  homogeneous features and the heterogeneous features of each modality from modality-aware (MA) prompt. The reconstructed modality feature embeddings demonstrate distributional similarity to ground-truth feature embeddings, validating the efficacy of our reconstruction module. From experimental observations, we find that when both Text and Image modalities are absent, relying solely on the Comment modality fails to reconstruct accurate feature vectors. This phenomenon originates from inherent modality characteristics: Text and images generally represent the content creator's personal, subjective perspective, whereas comments tend to express more objective perspective from external users such as reviewers or readers.

We conducted quantitative experiments to complement our qualitative visualizations and provided a more objective evaluation of the reconstructed modality features showed in Table \ref{tab:COS}. Specifically, we computed the cosine similarity between the reconstructed features and the ground truth (GT) features under various missing modality scenarios. As shown in the updated Table, the cosine similarity between the reconstructed and GT features ranges from 0.2765 to 0.6134 depending on the available modalities. In contrast, the cosine similarity between random features and GT features remains close to zero in all cases, confirming that our reconstruction is substantially better than random guessing. These quantitative results are consistent with our visualizations. This additional analysis demonstrates that our reconstruction method produces features that are quantitatively and qualitatively similar to the ground truth, and the results are not due to cherry-picking or visual bias.

\end{document}